%% file: deepseek_arxiv.tex
\documentclass{article}

     \usepackage[preprint]{neurips_2023}

\usepackage[utf8]{inputenc} 
\usepackage[T1]{fontenc}    
\usepackage{hyperref}       
\usepackage{url}            
\usepackage{booktabs}       
\usepackage{amsfonts}       
\usepackage{nicefrac}       
\usepackage{microtype}      
\usepackage{xcolor}         
\usepackage{amsmath, bm}
\usepackage{amssymb}
\usepackage{mathtools}
\usepackage{amsthm}
\usepackage{hyperref}
\usepackage{algorithm}
\usepackage{algorithmic}

\input{maths_commands}
\usepackage{subcaption}   

\usepackage{xcolor}
\usepackage{subcaption}
\usepackage{booktabs}
\usepackage{float}    
\usepackage{tcolorbox}
\usepackage{listings}
\lstdefinestyle{greenstyle}{
	language=Python,
	backgroundcolor=\color{green!10},      
	basicstyle=\ttfamily\small\color{green!50!black}, 
	frame=single,
	rulecolor=\color{green!80!black},       
	breaklines=true,
	numbers=left,                          
	numberstyle=\tiny\color{green!50!black},
	stepnumber=1,
	numbersep=10pt,
}

\title{HessFormer: Hessians at Foundation Scale}

\author{%
  Diego Granziol \\
  Maths Department\\
  University of Oxford\\
  \texttt{granziol@maths.ox.ac.uk} \\
}

\begin{document}

\maketitle

\begin{abstract}
Whilst there have been major advancements in the field of first order optimisation of deep learning models, where state of the art open source mixture of expert models go into the hundreds of billions of parameters, methods that rely on Hessian vector products, are still limited to run on a single GPU and thus cannot even work for models in the billion parameter range. We release a software package \textbf{HessFormer}, which integrates nicely with the well known Transformers package and allows for distributed hessian vector computation across a single node with multiple GPUs. Underpinning our implementation is a distributed stochastic lanczos quadrature algorithm, which we release for public consumption. Using this package we investigate the Hessian spectral density of the recent Deepseek $70$bn parameter model.
\end{abstract}
\section{Introduction}

Large-scale language models now routinely exceed $10^{11}$ trainable parameters, yet our theoretical understanding of their optimisation landscape has stagnated at the million-parameter scale where exact second-order information is still tractable.  
The \textit{Hessian} the matrix of second derivatives of the loss with respect to the parameters encodes the local curvature of that landscape.  
Its spectrum governs everything from the step-size limits of first-order optimisers \citep{cohen2021edge}, to the effectiveness of curvature-aware training algorithms such as Sophia \citep{liu2023sophia}, to the reliability of post-hoc interventions such as influence functions and unlearning \citep{grosse2023influence, jia2024soul}.  
Unfortunately, computing even a single Hessian vector product (HVP) for today’s 10--100~billion-parameter Transformers already exceeds the memory of a single GPU, and existing distributed frameworks deliberately detach gradient collectives from the autograd graph, rendering naive double-backward operations impossible \citep{huang2023fsdp}.  
As a result, \emph{no} publicly available work has reported the full Hessian spectrum of a model larger than 7B parameters.

We close this gap with \textbf{HessFormer}, the first software framework that enables \emph{distributed} stochastic Lanczos quadrature on PyTorch models without modifying user code or sacrificing data, tensor, or pipeline parallelism.  
Leveraging the observation that HVPs can be expressed as a sum of inexpensive local contractions once gradients are preserved in graph form, we design a lightweight communication protocol that performs these contractions across GPUs and overlaps them with forward passes.  
Running on a single 8\,$\times$\,A200 node, \textbf{HessFormer} estimates hundreds of leading Hessian eigenvalues of \textit{DeepSeek-LLM-70B} \cite{deepseek2025r1}.

\subsection*{Why does a 70\,B-parameter Hessian spectrum matter?}
\citet{dettmers2022llm} show that once a model exceeds $6$B parameters, a handful of hidden-state dimensions become “extreme outliers” (up to 20× larger than typical activations). These outliers concentrate most of the model’s predictive power, so the authors create a mixed-precision scheme that keeps just those features in 16-bit while quantising everything else to 8-bit.  \citet{sun2025phase} use an Ising model of the transformer to model phase transition, and find that at $7$B parameters, transformers manage to learn whether their nonsensical outputs are nonsensical or not and thus represent fundamentally different learning machines. Further use cases may include

\begin{enumerate}
	\item \textbf{Optimisation safety and speed.}  
	Tracking the extreme eigenvalues of $H$ lets practitioners set learning rates and trust-region radii on firm ground rather than heuristics.
	\item \textbf{Validation of curvature approximations.}  
	Block-diagonal, diagonal, and Kronecker-factor assumptions pervade second-order optimisers and pruning criteria.  
	Only empirical spectra at ``foundation'' scale reveal where those assumptions break down.
	\item \textbf{Diagnostics and monitoring.}  
	Spikes in the top eigenvalues often precede gradient explosions, while clusters of near-zero eigenvalues indicate flat directions and compression opportunities.
	\item \textbf{Influence, unlearning, and continual learning.}  
	All three require solves of the form $(H+\lambda I)^{-1}v$.  
	Fast HVPs move these techniques from toy models toward production-size LLMs.
	\item \textbf{Scientific insight.}  
	Scaling-law research has so far focused on loss and gradient norms; Hessian spectra add a missing dimension, informing theories of sharp-vs-flat minima, mode connectivity, and transferability.
\end{enumerate}

By turning HVPs into a first-class distributed primitive, \textbf{HessFormer} closes the tooling gap between million-parameter theory and billion-parameter practice and furnishes the empirical groundwork for the next generation of optimisation, interpretability, and safety methods.

\section{Motivation}

\begin{table}[h]
	\centering
	\renewcommand{\arraystretch}{1.2}
	\setlength{\tabcolsep}{6pt}
	\begin{tabular}{|c|c|c|c|}
		\hline
		\citet{KohLiang2017} &
		\citet{HassibiStork1993} &
		\citet{Nagel2020} &
		\citet{Martens2010} \\
		\hline
		\scriptsize\shortstack{
			$H_\theta=\frac1n\sum_{i}\nabla_\theta^{2}L(z_i,\theta)$\\[2pt]
			$\delta\theta=-\,H_{\hat\theta}^{-1}\nabla_\theta L(z_0,\hat\theta)$
		} &
		\scriptsize\shortstack{
			$\Delta w=-\,w_q/(H^{-1})_{qq}\,H^{-1}e_q$\\[2pt]
			$\Delta E=w_q^{2}\big/\!\bigl(2\,(H^{-1})_{qq}\bigr)$
		} &
		\scriptsize\shortstack{
			$\displaystyle
			\min_{\tilde w}\,
			\frac12\,(\tilde w-w)^{\!\top}H(\tilde w-w)$\\[2pt]
			$\text{s.t. }\tilde w_i\in
			\{\!\lfloor w_i\rfloor_{\text{g}},
			\lceil w_i\rceil_{\text{g}}\}\;\forall i$
		} &
		\scriptsize\shortstack{
			$\delta w=-\,H^{-1}\nabla_w\mathcal L(w)$\\[2pt]
			$w_{\text{new}}=w+\delta w$
		}\\
		\hline
	\end{tabular}
	\caption{Key Hessian-based formulae for four different machine-learning techniques. (a) Influence functions, (b) Optimal brain surgeon, (c) Adaptive rounding, (d)  Second order optimisation.}
	\label{tab:hessian_use_cases}
\end{table}

The Hessian of a neural network, i.e. the matrix of second-order partial derivatives of the loss with respect to the parameters, is far more than an academic curiosity. As Table \ref{tab:hessian_use_cases} shows, it is the central mathematical object behind four major research agendas: influence estimation \citep{KohLiang2017}, network surgery and pruning \citep{HassibiStork1993}, post-training quantisation \citep{Nagel2020}, and second-order optimisation \citep{Martens2010}. In every case the method either depends directly on an inverse Hessian vector product (HVP) or builds its key approximation from local curvature. A reliable Hessian spectrum therefore offers a common empirical benchmark against which techniques that tackle very different problems can be compared.
Despite this centrality, empirical knowledge of Hessian spectra stops at networks of about seven billion parameters. For smaller convolutional and transformer models, three patterns appear repeatedly in the literature: strong block-level heterogeneity of eigenvalues that explains why adaptive optimisers outperform SGD \citep{zhang2024transformersadam}; analytic cross-term structure unique to self-attention \citep{ormaniec2024transformerhessian}; and heavy-tailed eigenvalue decay that links deep learning to random-matrix theory \citep{xie2022powerlawhessian}. Whether any of these properties persist, flatten, or split in the 70 B to 1 T range is unknown. Curvature-aware methods that work well at modest scale, Hessian trace regularisers \citep{sankar2020hessianregularization}, diagonal optimisers such as Sophia \citep{liu2023sophia}, Kronecker factor approximations (K-FAC, EK-FAC), and projection-based influence estimation (LoGra) are therefore being applied to foundation models without direct validation.
\newline
\newline
The technical reason for this gap is simple. Computing a single HVP for a 70 B parameter transformer requires a second backward pass that doubles activation memory, stores an extra gradient-sized buffer, and, crucially, performs gradient collectives \emph{inside} the autograd graph. Popular sharding frameworks such as ZeRO and FSDP break this requirement: their gradient hooks run the reduce-scatter outside the graph and then set \texttt{p.grad = None}, severing every higher-order path \citep{huang2023fsdp}. Compiler systems like GSPMD \citep{xu2021gspmd} could in principle restore differentiability, but no public PyTorch stack exposes that capability.
The absence of ground-truth curvature data between 10 B and 100 B parameters has practical consequences. We do not know whether the heavy tails seen at smaller scales become plateaus at mixture-of-experts boundaries, how far diagonal, block-diagonal, or Kronecker assumptions drift from reality when width and depth grow together, or whether the instabilities observed during billion-scale fine-tuning come from optimiser settings or genuinely extreme curvature. Accurate Hessian access is also essential for trustworthy unlearning \citep{jia2024soul}, data valuation and back-door detection \citep{zou2023repsim}, and structured compression techniques such as WoodFisher and Optimal BERT Surgeon \citep{singh2020woodfisher,kurtic2022obert}.
Addressing these unknowns requires a tool that reveals second-order structure without dismantling the memory-efficient training stacks on which modern language models rely. By enabling distributed stochastic Lanczos quadrature for DeepSeek-LLM-70B on a single eight-GPU node, \textsc{HessFormer} fills this methodological void. The spectra it produces provide the first concrete evidence of curvature at foundation-model scale, allowing the community to calibrate optimisation, compression, and safety techniques against reality rather than conjecture.

\section{Distributed Pealmutter \& Stochastic Lanczos Quadrature}

\subsection{Huggingface Auto}
\begin{figure}[ht]
	\centering
	\begin{subfigure}[b]{0.34\textwidth}
		\includegraphics[width=\textwidth]{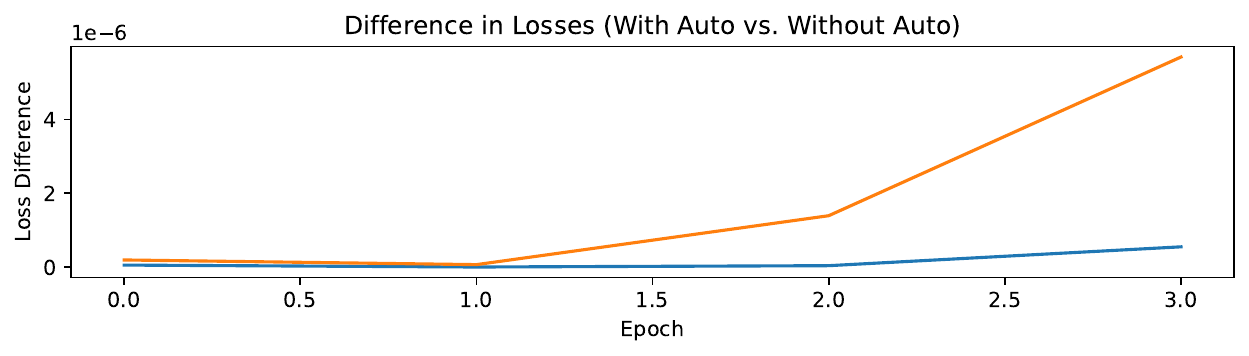}
		\caption{Loss Difference magnified}
		\label{fig:comparisonauto:losses}
	\end{subfigure}
	\begin{subfigure}[b]{0.3\textwidth}
		\includegraphics[width=\textwidth]{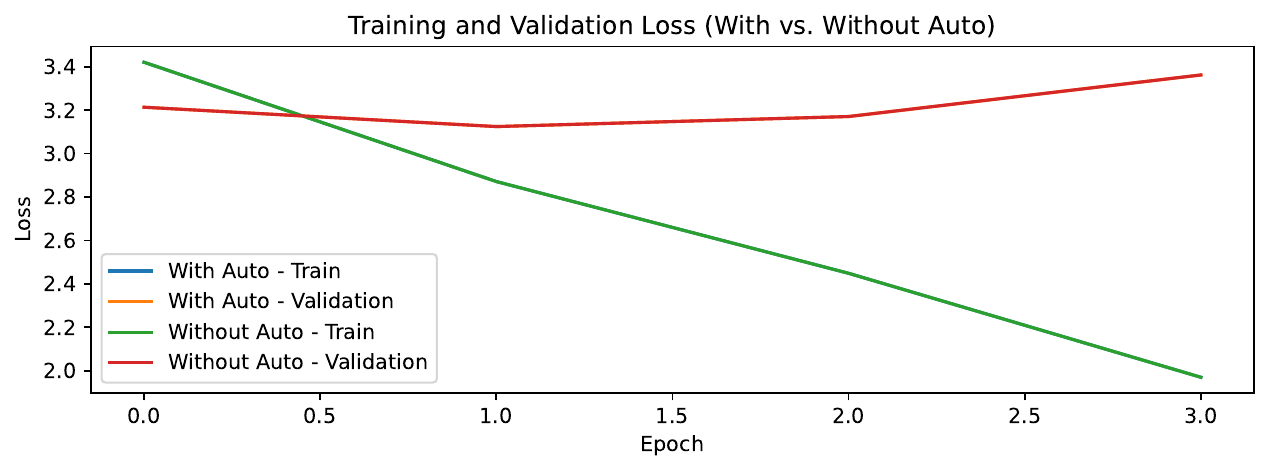}
		\caption{Loss trajectory}
		\label{fig:comparisonauto:train}
	\end{subfigure}
	\begin{subfigure}{0.34\textwidth}
		\includegraphics[width=\textwidth]{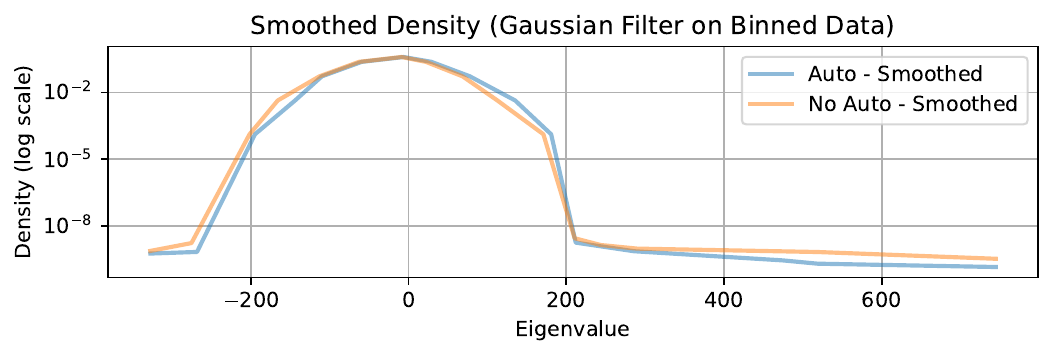}
		\caption{spectral density difference}
		\label{fig:smoothed_density}
	\end{subfigure}
	\caption{Comparison between Training a $1.5$Bn parameter Qwen distilled Deepseek model on wikitrain for $4$ epochs using auto from huggingface on multiple GPUs against using a single large enough GPU training on cuda.}
	\label{fig:comparisonauto}
\end{figure}

We leverage the ubiqitous Huggingface ecosystem, which has its own distributed inference environment called \textit{"auto"}. One can specify which layers or part thereof go onto which GPU manually, but in this case instead of writing \textit{device="cuda:0"} we simply use their integrated solution \textit{device="auto"}.  As this feature is inteded for inference only and does not support training. We trial two small $1.5$bn parameter Deepseek with further training on next token prediction using the huggingface wikitext dataset \cite{merity2016pointer}. Both experiments run with identitcal hyperparameters, but for the baseline we use a $H\times 200 = 140$GB GPU instance and the other used $4 \times RTX 4094 = 4 \times 24$ GB multi GPU single node cluster. We compare the difference in training and validation loss results in Figures \ref{fig:comparisonauto:losses} and \ref{fig:comparisonauto:train}. We see differences at the $10^{-6}$ level, essentially GPU machine precision.  Notice how the lines completely overlap on the typical training plot. This indicates that the gradients are none the less calculated effectively using auto.
As a precursor, to the next section where we 
We further plot the differences in spectra when using auto or not. As shown in Figure \ref{fig:smoothed_density}, for a smoothed density, we see barely perceptible difference. Note that we are only using a single $n=1$ seed vector. So we expect different results between densities using different seed vectors. Even though we fix the random vector and init to be $seed=42$ we do expect differences machine to machine.


\paragraph{Inefficiency:}
\label{sec:fsdp}
One of the reasons why training LLMs with device auto is not standard is that it places (some subset of/each) layer(s) on idividual GPUs. What this means is that at any one point $n/n-1$ GPUs will be idle, inefficient in the large $n$ limit. A better method, known as fully sharded data parallel is more efficient, but as detailed in the introduction does not allow for double backward pass.
\subsection{DSLQ}
Unlike fully sharded data parallel implementations, the auto layerwise splitting does not automatically destroy the graph for memory efficiency and so it is possible to proceed with a double backward pass (otherwise known as Pearlmutters trick \citet{pearlmutter1994fast}), which we detail in Pseudocode in Algorithm \ref{alg:hvp_multi}. We combine this with distributed stochastic lanczos quadrature without orthogonalisation, which we detail in Algorithm \ref{alg:lanczos_chunk}. 

\begin{figure}[ht]
	\centering
	\begin{subfigure}[b]{0.32\textwidth}
		\includegraphics[width=\textwidth]{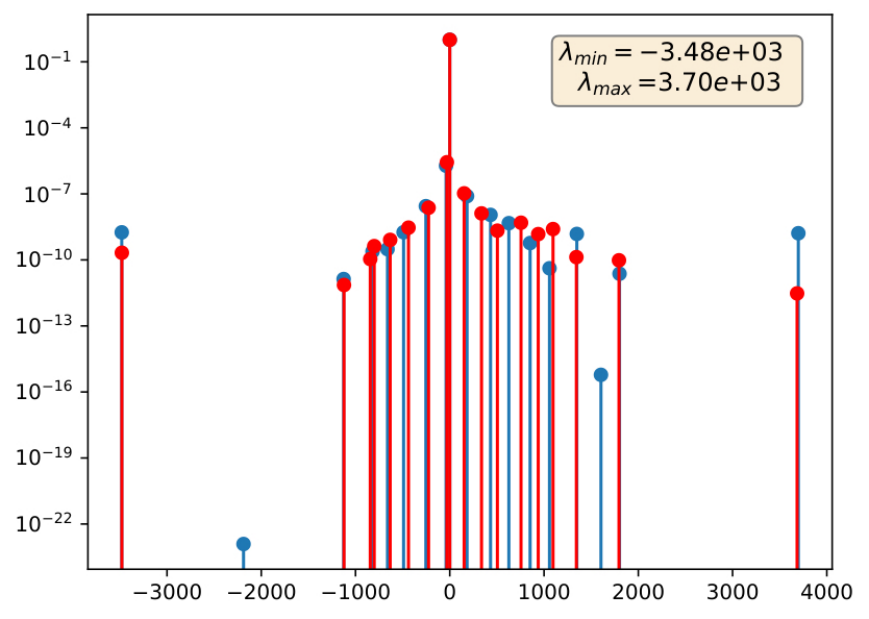}
		\caption{Random Dataset}
		\label{subfig:randomautovsnot}
	\end{subfigure}
	\hfill
	\begin{subfigure}[b]{0.32\textwidth}
		\includegraphics[width=\textwidth]{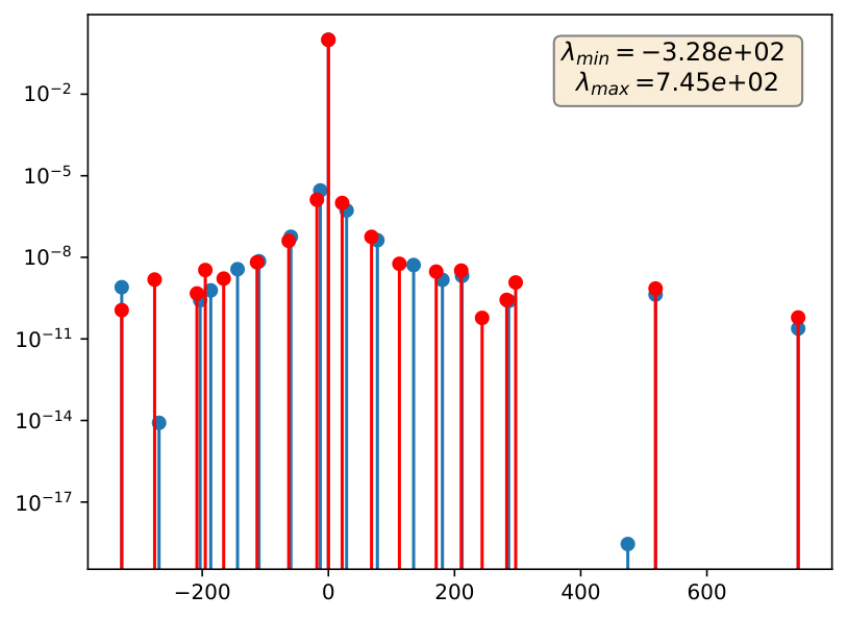}
		\caption{Wikitext}
		\label{subfig:wikiautovsnot}
	\end{subfigure}
	\hfill
	\begin{subfigure}[b]{0.32\textwidth}
		\includegraphics[width=\textwidth]{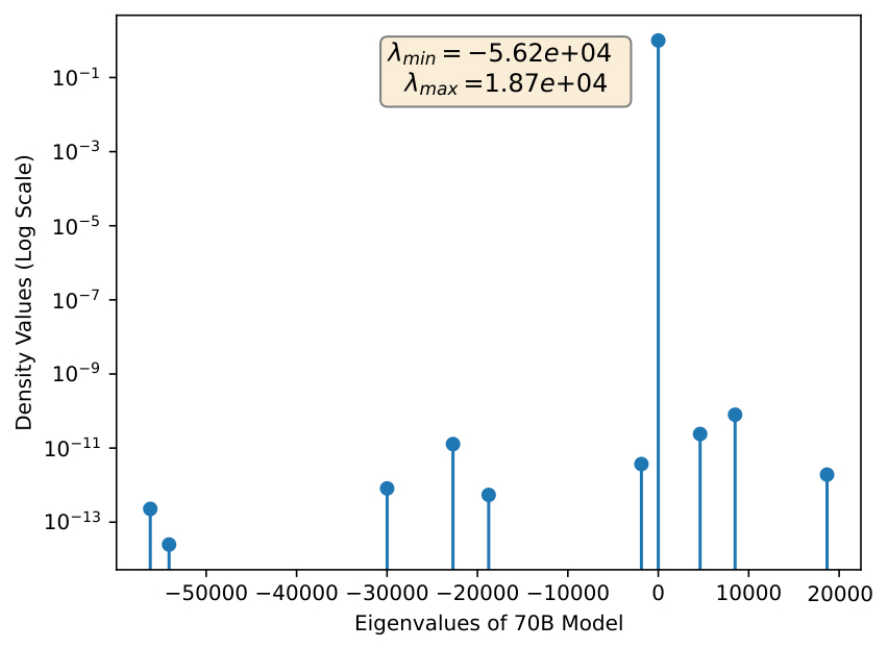}
		\caption{Pile (70B, 10 10 \%)}
		\label{subfig:pile70b}
	\end{subfigure}
	
	\caption{Stem comparison of using Auto (blue) vs.\ not using Auto (red) for Hessian of the DeepSeek-distilled Qwen models across three datasets. The rightmost subfigure is an example of how a large model looks on a common dataset.}
	\label{fig:stemautovsnotrand}
\end{figure}

We conduct ablation experiments to check the validity of our framework in preserving crucial spectral information. We test for a small $1.5$bn parameter model, how using our distributed setting alters the spectral resolution using a single random vector. As seen in Figure \ref{fig:stemautovsnotrand}, plotting in auto blue and the single GPU in red, the stem plots are basically identical. The only thing we see substantially difference is this one time outlier value for the wikitext dataset with a weight of $10^{-27}$. This is significantly smaller than the estimation of a single eigenvalue on the entire matrix and so represents with high probability a "ghost" from the Lanczos procedure due to no orthogonalisation. 

\subsection*{Memory Compute trade off: Investigating the Ghosts}
\begin{figure}[h!]
	\centering
	\begin{subfigure}[b]{0.24\textwidth}
		\includegraphics[width=\textwidth]{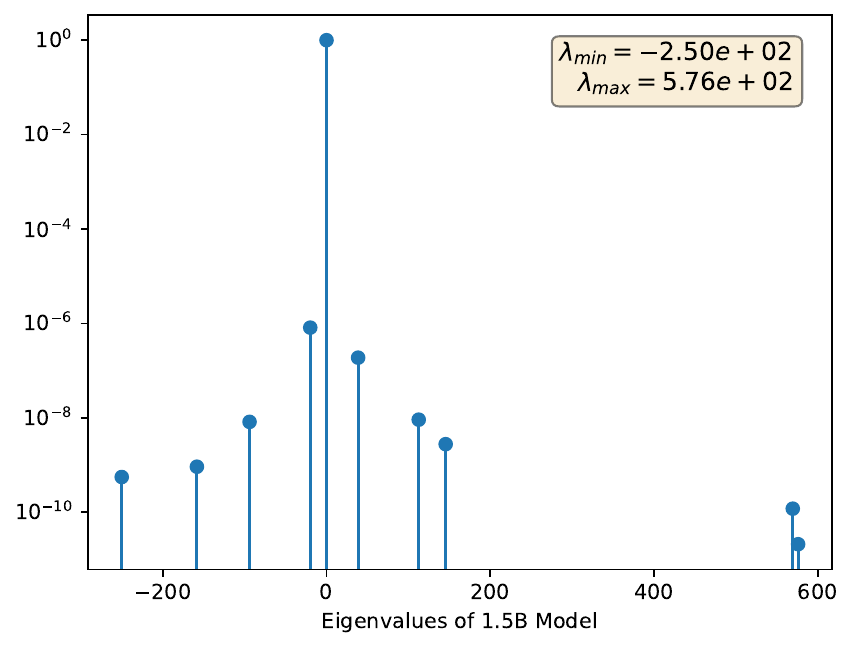}
		\caption{$100\%$\,–\,10 iters}
		\label{fig:percent:100}
	\end{subfigure}
	\hfill
	\begin{subfigure}[b]{0.24\textwidth}
		\includegraphics[width=\textwidth]{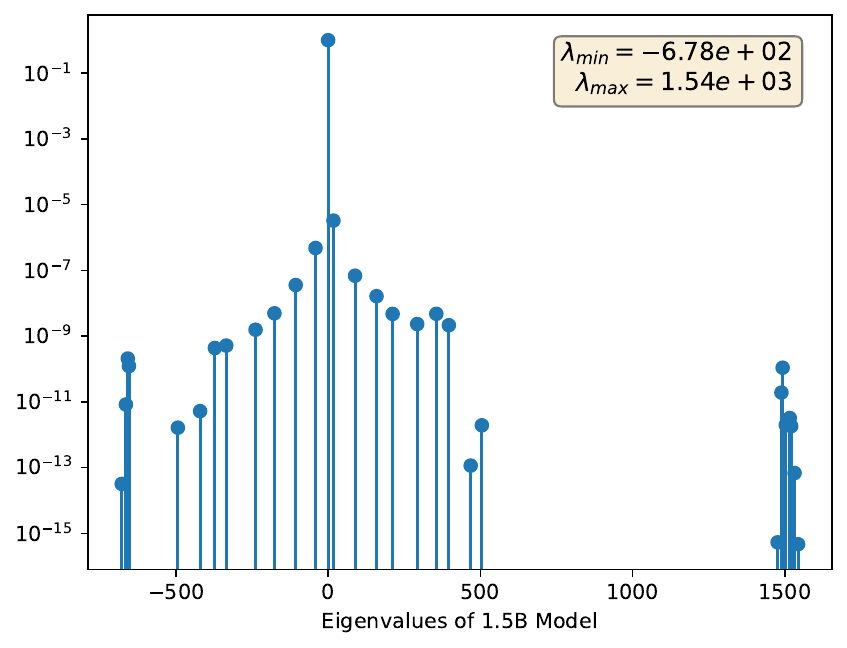}
		\caption{$1\%$\,–\,30 iters}
		\label{fig:percent:1}
	\end{subfigure}
	\hfill
	\begin{subfigure}[b]{0.24\textwidth}
		\includegraphics[width=\textwidth]{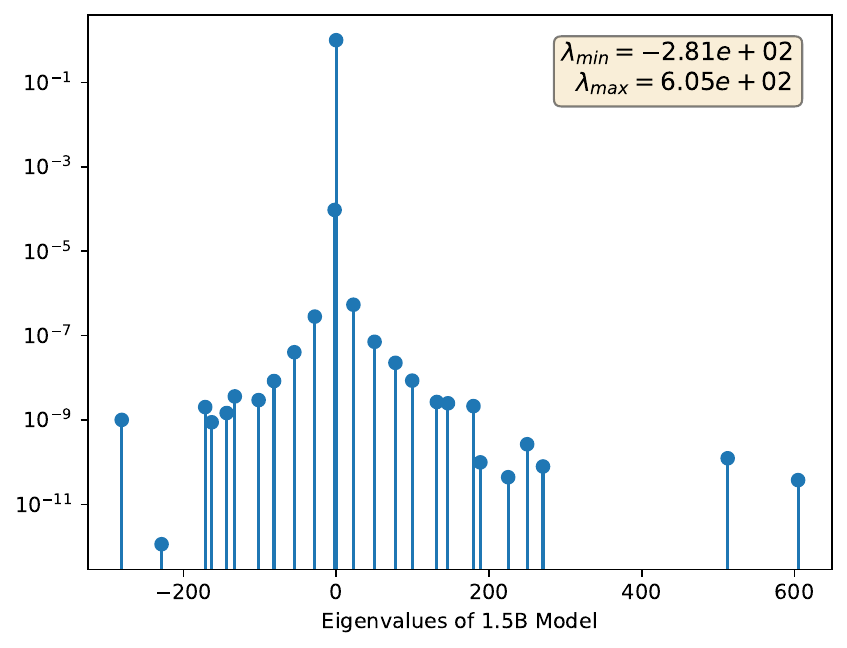}
		\caption{Full Ortho}
		\label{fig:fullortho}
	\end{subfigure}
	\hfill
	\begin{subfigure}[b]{0.24\textwidth}
		\includegraphics[width=\textwidth]{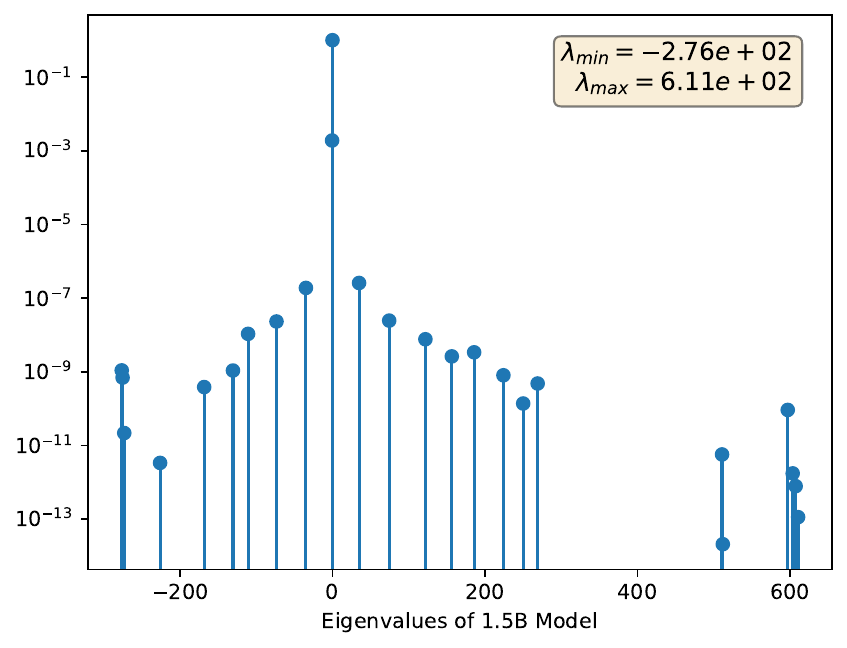}
		\caption{No Ortho}
		\label{fig:noortho}
	\end{subfigure}
	
	\caption{DeepSeek Qwen 1.5 Bn: spectral effects of data subsampling and orthogonalization on the wikitext dataset.}
	\label{fig:combined_wiki_ortho}
\end{figure}

One may note in Figures \ref{fig:percent:100}, \ref{fig:percent:1} as well as in the previous Figure \ref{fig:stemautovsnotrand} that both plots have a bunching of eigenvalues at the right end by the largest magnitude eigenvalue.  Since covariance matrix decomposition from linear regression into the within-cluster and between-cluster scatter
\[
\underbrace{\sum_{\ell=1}^k \sum_{i=1}^{n_\ell} (x_{\ell,i} - \bar{x}_\ell)(x_{\ell,i} - \bar{x}_\ell)^\top}_{\text{Within-Cluster Scatter}}
\;+\;
\underbrace{\sum_{\ell=1}^k n_\ell \bigl(\bar{x}_\ell - \bar{x}\bigr)\bigl(\bar{x}_\ell - \bar{x}\bigr)^\top}_{\text{Between-Cluster Scatter}},
\]
would suggest there is a singular mean eigenvalue of largest size, we investigate whether this is a linear algebra artifact.
We run the $1.5$bn parameter Qwen distil with two variants. For the baseline we use full orthonormalisation by storing the $Q$ matrix in Lanczos and for the trial we simply keep the three term recurrence relation and pretend we are in infinite precision. As can be seen in Figure \ref{fig:noortho}, whilst we keep the general information in terms of the spectral support and shape of the bulk, we start to get \emph{ghost} eigenvalues, against the baseline in Figure \ref{fig:fullortho}. This is nothing more than the memory complexity trade off but for stochastic Lanczos quadrature. By not enforcing orthogonality (which is impractical for large networks due to the extra memory cost) we end up repeating (essentially wasting computation). In this Figure we have only used $25$ iterations and we already have multiple ghosts, as such we believe that it is prudent to keep the number of iterations less than $20$ and for extremely large experiments we often use the value $10$ to not waste computation in the non orthogonal finite precision paradigm.

\begin{figure}[htb]
	\centering
	\begin{minipage}[t]{0.48\textwidth}
		\scriptsize
		\begin{algorithm}[H]
			\caption{Hessian–Vector Product (multi-GPU, double backward)}
			\label{alg:hvp_multi}
			\begin{algorithmic}[1]
				\STATE \textbf{Input:} model $\mathcal{M}$,\; loader $\mathcal{D}$,\; vectors $\{v_i\}$
				\STATE $v_i \leftarrow v_i.\texttt{to}(\mathrm{device}(p_i))$
				\STATE $h_i\gets0$, $N\gets0$, $D_{\text{main}}\gets\mathrm{device}(p_1)$
				\STATE $\mathcal{M}.\texttt{eval}()$
				\FOR{\textbf{each} batch $B$ in $\mathcal{D}$}
				\STATE $B \leftarrow B.\texttt{to}(D_{\text{main}})$
				\STATE $\mathcal{M}.\texttt{zero\_grad}(\texttt{set\_to\_none}= \texttt{True})$
				\STATE \textbf{with} \texttt{sdp\_kernel}(flash$=$False):
				\STATE \hspace{0.7em}$L \leftarrow \mathcal{M}(B).\texttt{loss}$
				\STATE \hspace{0.7em}$g_i \leftarrow \partial L/ \partial p_i$ \hfill(create\_graph)
				\STATE \hspace{0.7em}$d \leftarrow \sum_i\!\bigl((g_i v_i).\texttt{sum()}\bigr).\texttt{to}(D_{\text{main}})$
				\STATE $u_i \leftarrow \partial d/ \partial p_i$ \hfill(detach)
				\STATE $b \gets \lvert B\rvert$
				\STATE $h_i \gets h_i + u_i\,b$
				\STATE $N \gets N + b$
				\ENDFOR
				\STATE \textbf{return} $h_i / N \quad\forall i$
			\end{algorithmic}
		\end{algorithm}
	\end{minipage}
	\hfill
	\begin{minipage}[t]{0.48\textwidth}
		\scriptsize
		\begin{algorithm}[H]
			\caption{Lanczos Iteration (chunked, no re-orthog.)}
			\label{alg:lanczos_chunk}
			\begin{algorithmic}[1]
				\STATE \textbf{Input:} HVP $\mathcal{H}$, chunked params, $k_{\max}$, tol $\varepsilon$
				\STATE Draw random $q_0$;\; scatter to chunks;\; $q_0 \gets q_0/\lVert q_0\rVert$
				\FOR{$k = 0$ \textbf{to} $k_{\max}-1$}
				\STATE Gather $q_k$; $r_k \gets \mathcal{H}(q_k)$
				\IF{$k > 0$}
				\STATE $r_k \gets r_k - \beta_{k-1} q_{k-1}$
				\ENDIF
				\STATE $\alpha_k \gets \langle q_k, r_k\rangle$
				\STATE $r_k \gets r_k - \alpha_k q_k$
				\STATE $\beta_k \gets \lVert r_k\rVert$
				\IF{$\beta_k < \varepsilon$}
				\STATE \textbf{break}
				\ENDIF
				\STATE $q_{k+1} \gets r_k / \beta_k$
				\ENDFOR
			\end{algorithmic}
		\end{algorithm}
	\end{minipage}
\end{figure}

\subsection{Understanding the Dataset Impact}

\begin{figure}[ht]
	\centering
	\begin{subfigure}[b]{0.32\textwidth}
		\includegraphics[width=\textwidth]{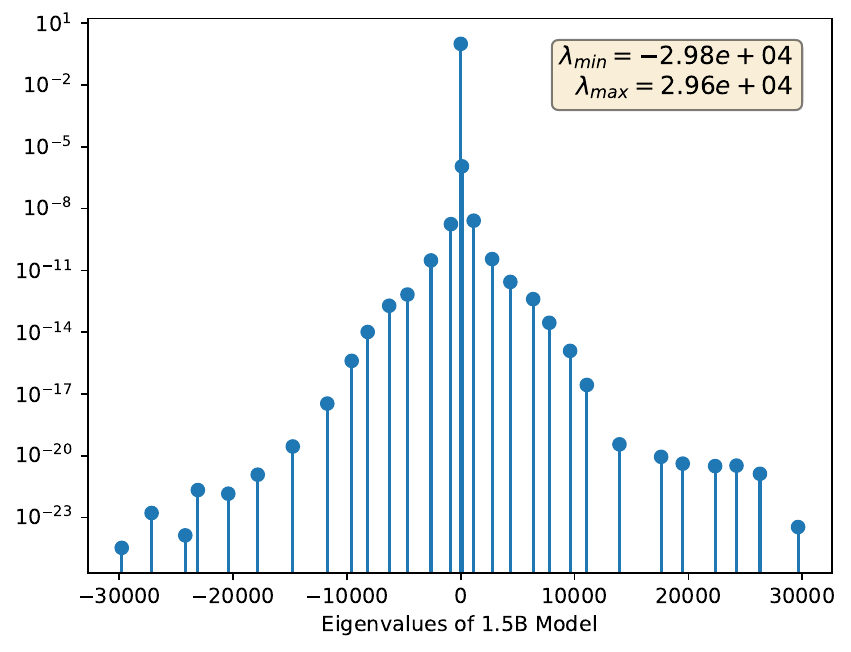}
		\caption{n=10000,l=64}
		\label{fig:randomdata:10000-64}
	\end{subfigure}
	\hfill
	\begin{subfigure}[b]{0.32\textwidth}
		\includegraphics[width=\textwidth]{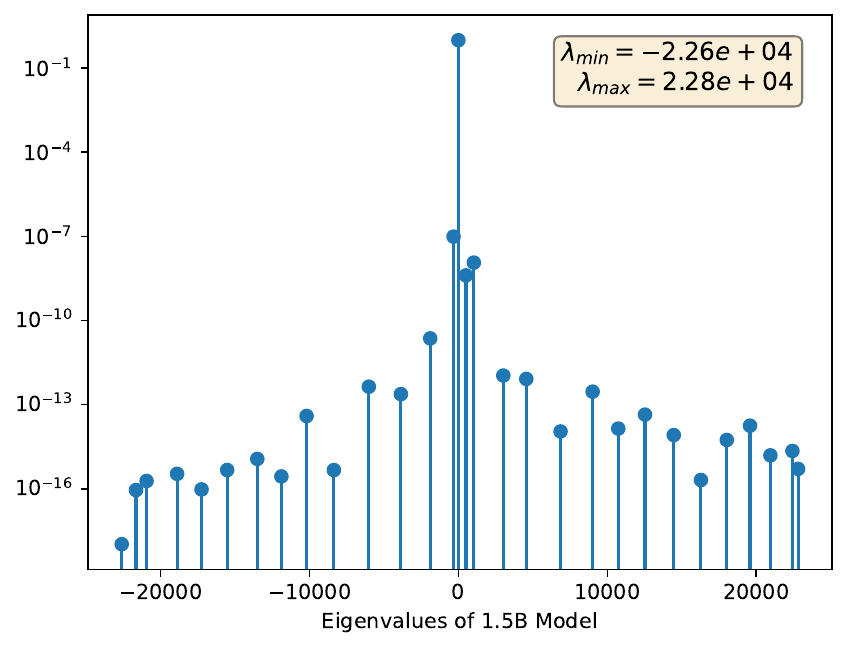}
		\caption{n=1000,l=1024}
		\label{fig:randomdata:1000-1024}
	\end{subfigure}
	\hfill
	\begin{subfigure}[b]{0.32\textwidth}
		\includegraphics[width=\textwidth]{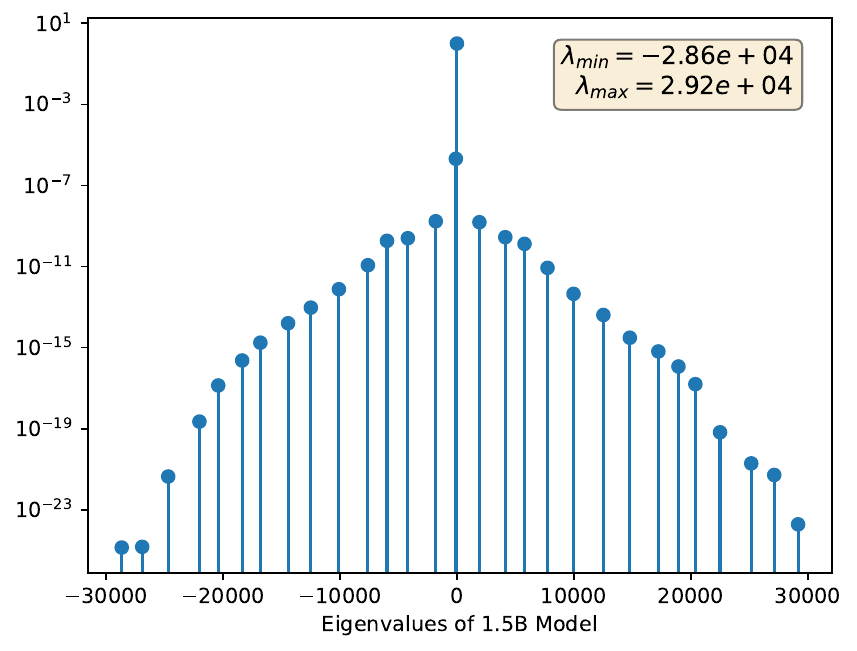}
		\caption{n=1000,l=64}
		\label{fig:randomdata:1000-64}
	\end{subfigure}
	\caption{Random Data with Various (n=num\_samples, l=seq\_length) Hessian with $30$ Lanczos HVP iterations.}
	\label{fig:randomdata-group}
\end{figure}

As shown in Figure \ref{fig:randomdata-group}. For random data, we see a reasonably consistent spectral density (with large near zero peak, no outliers, compact support) for various choices of sample size and sequence length on a random dataset on a trained LLM. Interestingly for Figure \ref{fig:randomdata:1000-64}, we see what looks like a degenrate semi cricle.

\begin{figure}[ht]
	\centering
	\begin{subfigure}[b]{0.32\textwidth}
		\includegraphics[width=\textwidth]{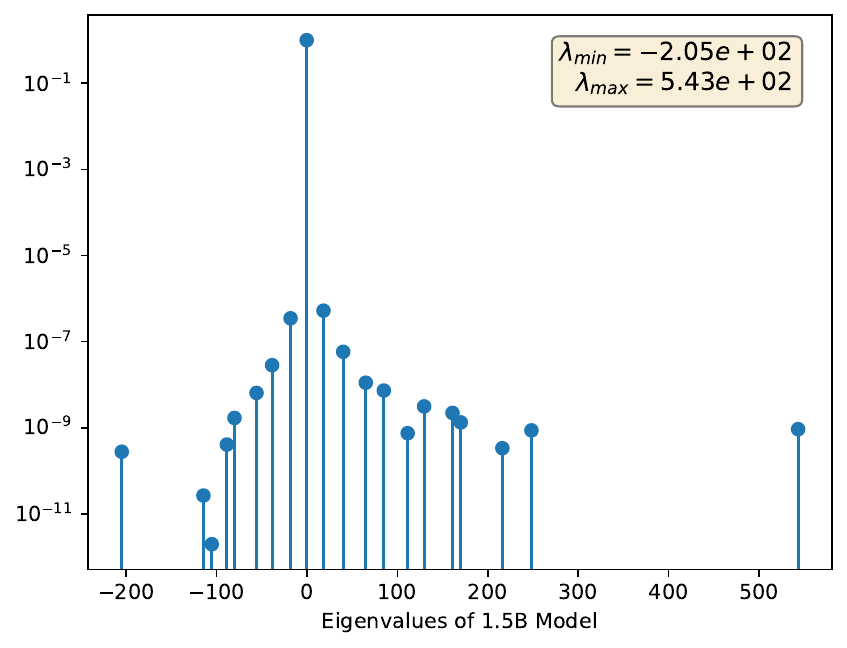}
		\caption{$1 \%$}
		\label{fig:realnews:0p01}
	\end{subfigure}
	\hfill
	\begin{subfigure}[b]{0.32\textwidth}
		\includegraphics[width=\textwidth]{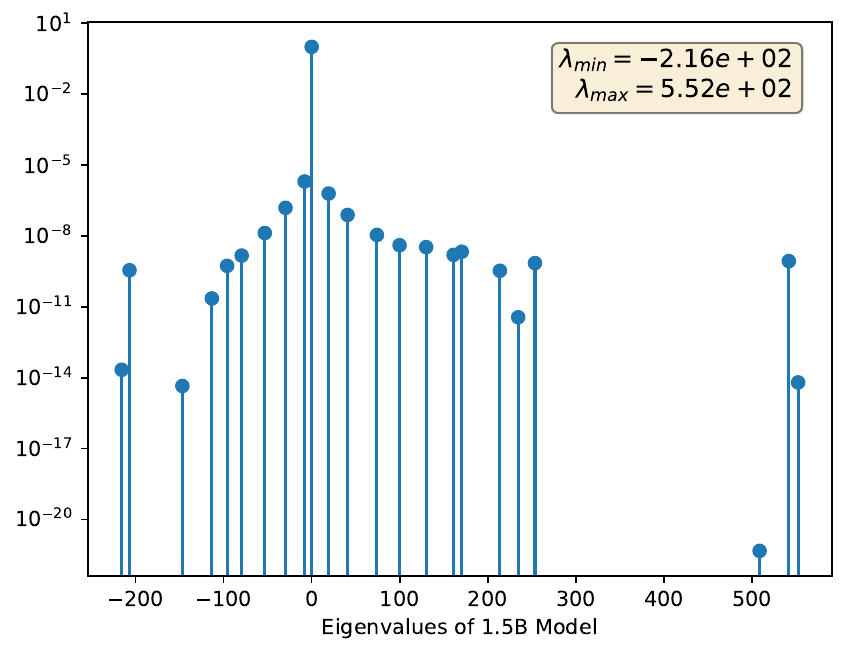}
		\caption{$10 \%$}
		\label{fig:realnews:0p1}
	\end{subfigure}
	\hfill
	\begin{subfigure}[b]{0.32\textwidth}
		\includegraphics[width=\textwidth]{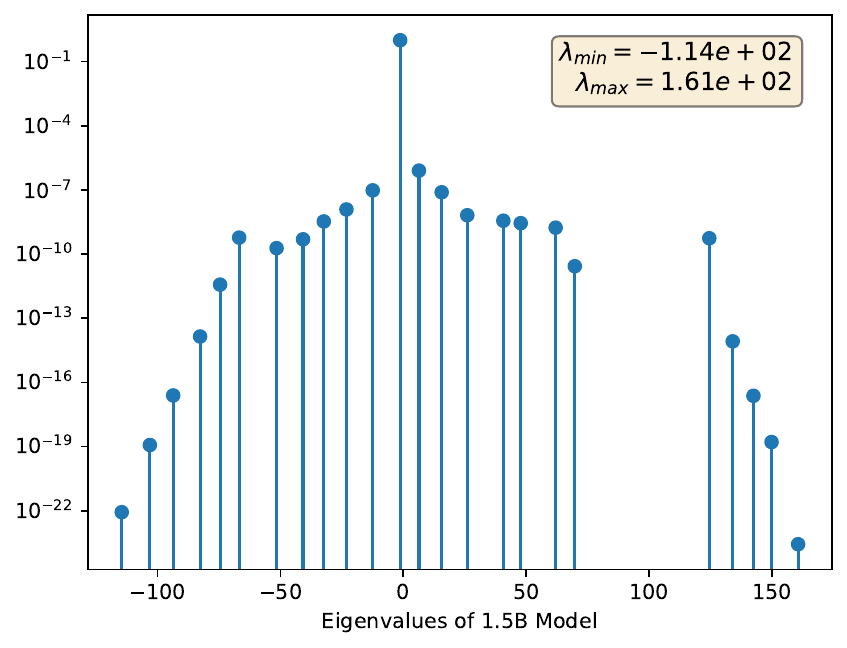}
		\caption{$100 \%$}
		\label{fig:realnews:1}
	\end{subfigure}
	\caption{Real News Data (subsampled to $1\%$) with various amounts of further subsampling Hessia for Qwen $1.5$Bn parameter DeepSeek distilled}
	\label{fig:realnews-group}
\end{figure}

\noindent For the real news dataset \cite{zeiler2012adadelta} in Figure \ref{fig:realnews-group}, initially subsampled to $1\%$ of its total and \textit{further} subsample in varying amounts, we see a very different shape with power law type decay. Furthermore note the presence of outliers, of which there are more as we increase the dataset size. This is in strong contrast to classical vision models such as in \cite{granziollearning} in which the number out outliers does not hugely change as we subsample.
%

\begin{figure}[h!]
	\centering
	\begin{subfigure}[b]{0.24\textwidth}
		\includegraphics[width=\textwidth]{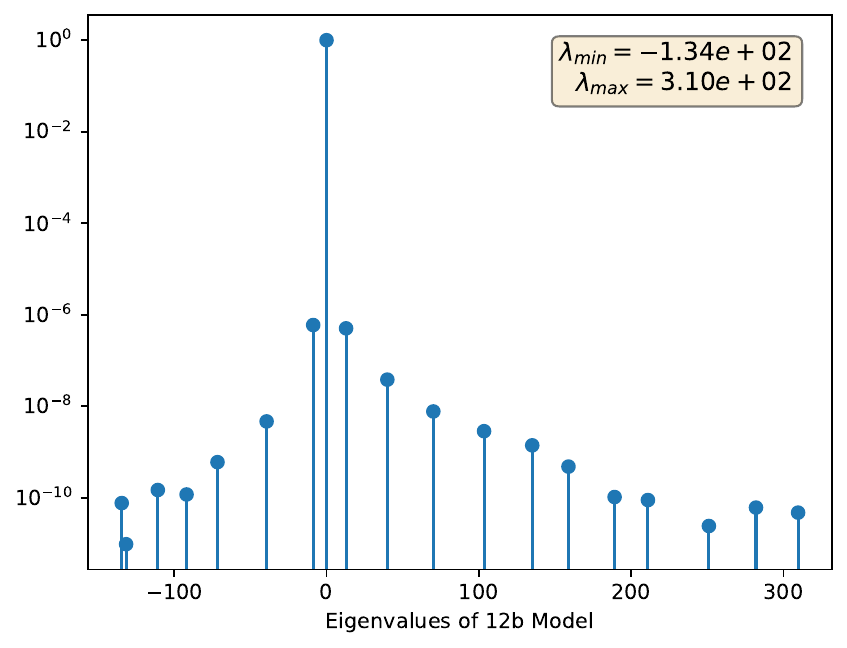}
		\caption{1\% Pile}
		\label{fig:pythia_12b_20it_1pc_pile}
	\end{subfigure}
	\hfill
	\begin{subfigure}[b]{0.24\textwidth}
		\includegraphics[width=\textwidth]{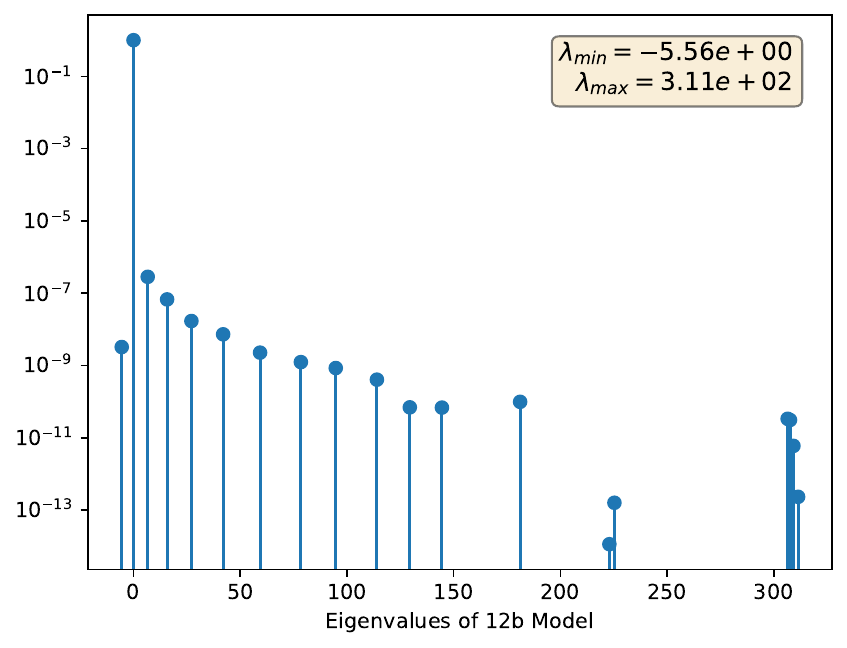}
		\caption{100\% Pile}
		\label{fig:pythia_12b_20it_100pc_pile}
	\end{subfigure}
	\hfill
	\begin{subfigure}[b]{0.24\textwidth}
		\includegraphics[width=\textwidth]{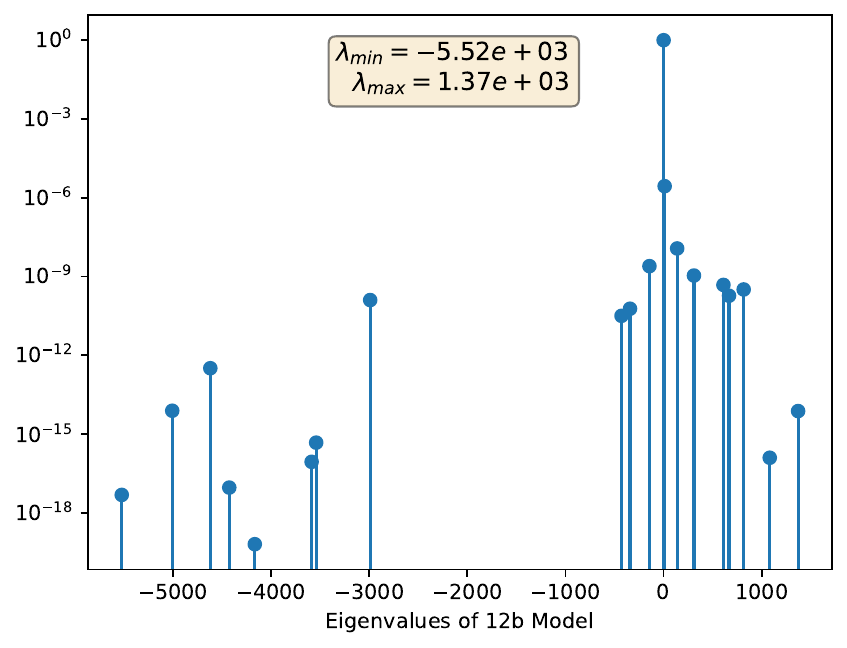}
		\caption{1\% Wiki}
		\label{fig:pythia_12b_20it_1pc_wiki}
	\end{subfigure}
	\hfill
	\begin{subfigure}[b]{0.24\textwidth}
		\includegraphics[width=\textwidth]{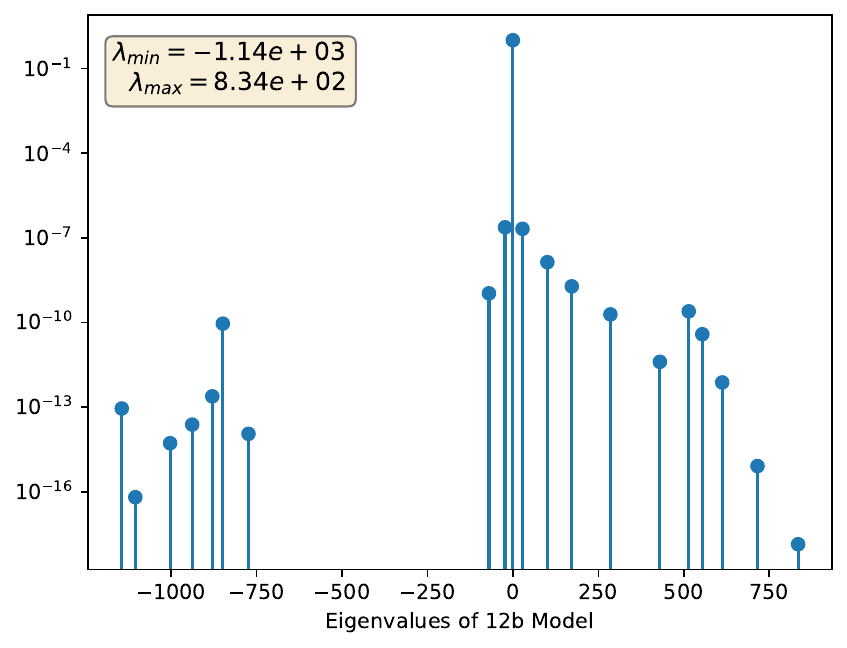}
		\caption{100\% Wiki}
		\label{fig:pythia_12b_20it_100pc_wiki}
	\end{subfigure}
	
	\caption{Pythia 12B (20 Lanczos iterations) spectral widths for the Pile and Wikipedia subsets.  
		The 10 \% Pile condition is omitted.}
	\label{fig:pythia_12b_20it_combined}
\end{figure}

We consider a case where we do know the data that the model was trained on. The Pythia \cite{biderman2023pythia} models trained on the $800$GB Pile dataset \cite{gao2020pile}. We see going from Figure \ref{fig:pythia_12b_20it_1pc_pile} to \ref{fig:pythia_12b_20it_100pc_pile}, that with less aggressive subsampling, the negative eigenvalues dissapear. However for an alternative dataset on which the model was not trained, as shown in FIgures \ref{fig:pythia_12b_20it_1pc_wiki}, \ref{fig:pythia_12b_20it_100pc_wiki} increasing the dataset size, whilst it does reduce the magnitude of all eigenvalues (as expected from broadening theorems) it does not reduce the large negative eigenvalues found here. 
\subsection{A note on numerical precision for the weight value}
In IEEE-754 float32 arithmetic we adopt \citet{higham2002accuracy} model \( \operatorname{fl}(a\!\circ\!b) = (a\!\circ\!b)\bigl(1+\delta\bigr) \) with unit-roundoff \( u = 2^{-24}\approx 5.96\times10^{-8} \) and accumulated error \( \gamma_{n}=nu/(1-nu) \) when \( nu\ll1 \).
The symmetric Lanczos recurrence computed without re-orthogonalisation satisfies
\[
\beta_{j}\widehat q_{j+1}
=A\widehat q_{j}-\alpha_{j}\widehat q_{j}
-\beta_{j-1}\widehat q_{j-1}+e_{j},
\qquad
\|e_{j}\|
\le\bigl(\|A\|\gamma_{3}+\gamma_{2}\bigr)
\max\{\|\widehat q_{j}\|,\|\widehat q_{j-1}\|\}.
\]
\citet{paige1976error} analysis gives the loss-of-orthogonality estimate
\[
\lvert\widehat q_{i}^{\mathsf T}\widehat q_{j}-\delta_{ij}\rvert
\le k\,u+{\cal O}(u^{2}),
\]
so each column of \( \widehat Q_{k} \) is perturbed by \( {\cal O}(ku) \).
After \( k=10 \) steps we solve \( T_{k}\widehat y=\theta\widehat y \) exactly and form the Ritz vector \( \widehat v=\widehat Q_{k}\widehat y \) whose components are
\( \widehat v_{i}=\sum_{j=1}^{k}\widehat Q_{ij}\widehat y_{j} \).
One inner product introduces at most \( \gamma_{k-1}\lvert\widehat v_{i}\rvert \) error, and basis perturbations add \( {\cal O}(ku)\lvert v_{i}\rvert \), giving
\[
\lvert\widehat v_{i}-v_{i}\rvert\le k\,u\,\lvert v_{i}\rvert.
\]
Squaring in float32 produces
\[
\widehat w_{i}=\operatorname{fl}\bigl(\widehat v_{i}^{2}\bigr)
=\widehat v_{i}^{2}(1+\delta_{\text{sq}}),
\qquad
\lvert\delta_{\text{sq}}\rvert\le u,
\]
and with \( \widehat v_{i}=v_{i}(1+\varepsilon_{i}) \) and \( \lvert\varepsilon_{i}\rvert\le k\,u \) we obtain
\[
\frac{\lvert\widehat w_{i}-w_{i}\rvert}{w_{i}}
=\bigl|(1+\varepsilon_{i})^{2}(1+\delta_{\text{sq}})-1\bigr|
\le 2\,k\,u + {\cal O}\!\bigl((ku)^{2}\bigr).
\]
For \( k=10 \) this bound is \( 2ku \approx 1.19\times10^{-6} \), so each squared Ritz-vector element retains roughly six accurate decimal digits in float32 before higher-order terms dominate. As such whilst it may make sense to introduce such a $10^{-6}$ noise floor, we find many of the bounds in these classic derivations simply to be too conservative. Note that many outliers have a roughly correct scaling (1/model param size). We are planning to release a random matrix theory derivation of the Lanczos error in upcoming work. As such we make an artistic decision to not touch the numerical plots.

\section{Frontier Spectra}
\begin{figure}[h!]
	\centering
	\begin{subfigure}[b]{0.23\textwidth}
		\includegraphics[width=\textwidth]{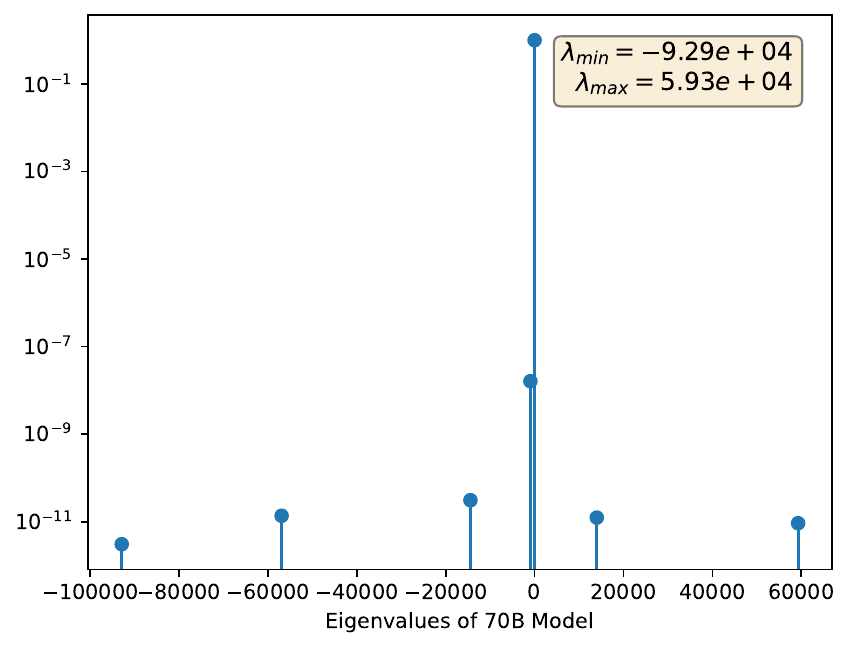}
		\caption{$5\%$ Wikitext, 10  It}
		\label{fig:70b_5pc_wikit}
	\end{subfigure}
	\hfill
	\begin{subfigure}[b]{0.23\textwidth}
		\includegraphics[width=\textwidth]{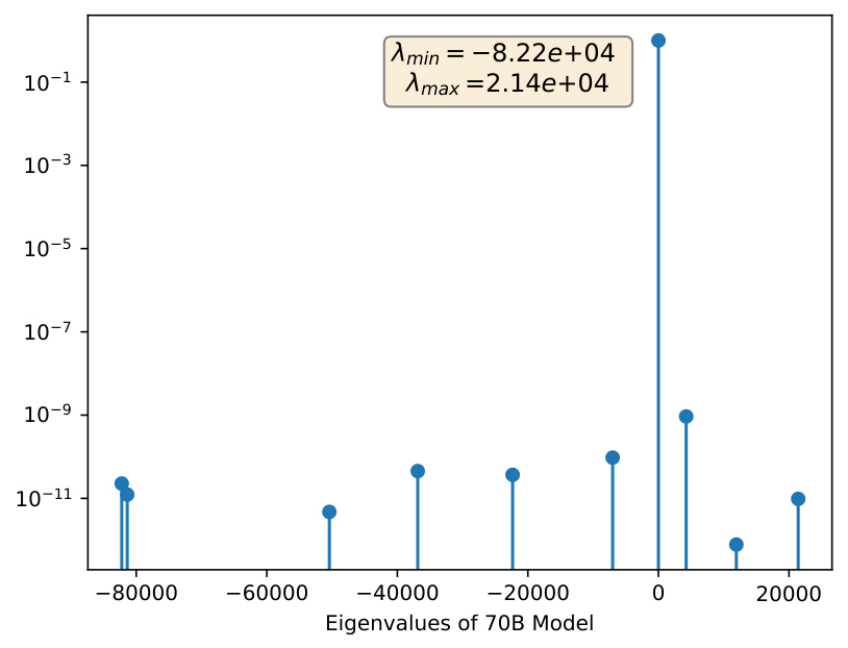}
		\caption{$25\%$ Wikitext, 10  It}
		\label{fig:70b_25pc_wiki}
	\end{subfigure}
	\hfill
	\begin{subfigure}[b]{0.23\textwidth}
		\includegraphics[width=\textwidth]{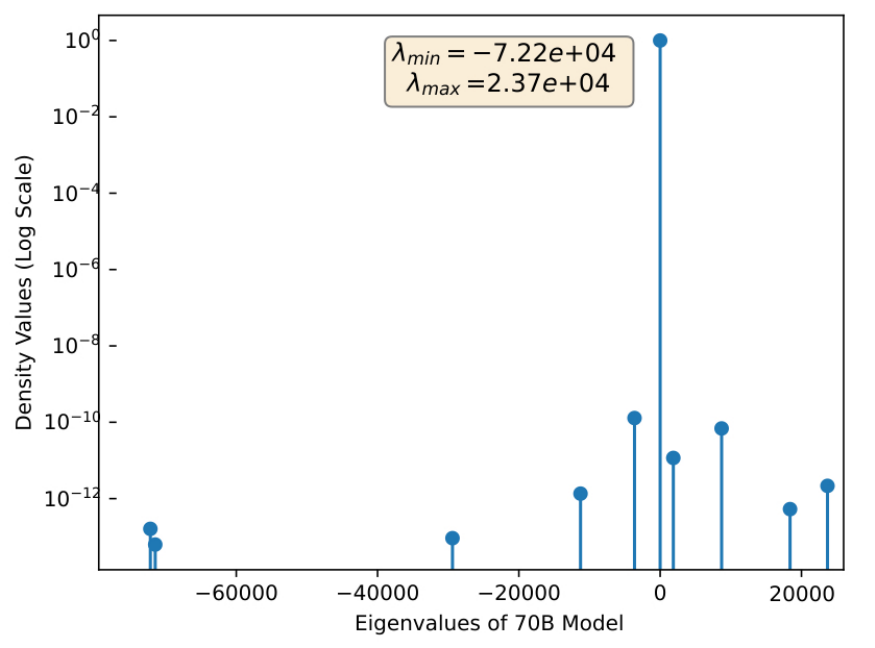}
		\caption{$0.1\%$ RealNews, 10  It}
		\label{fig:70b_0p1pc_10it}
	\end{subfigure}
	\hfill
	\begin{subfigure}[b]{0.23\textwidth}
		\includegraphics[width=\textwidth]{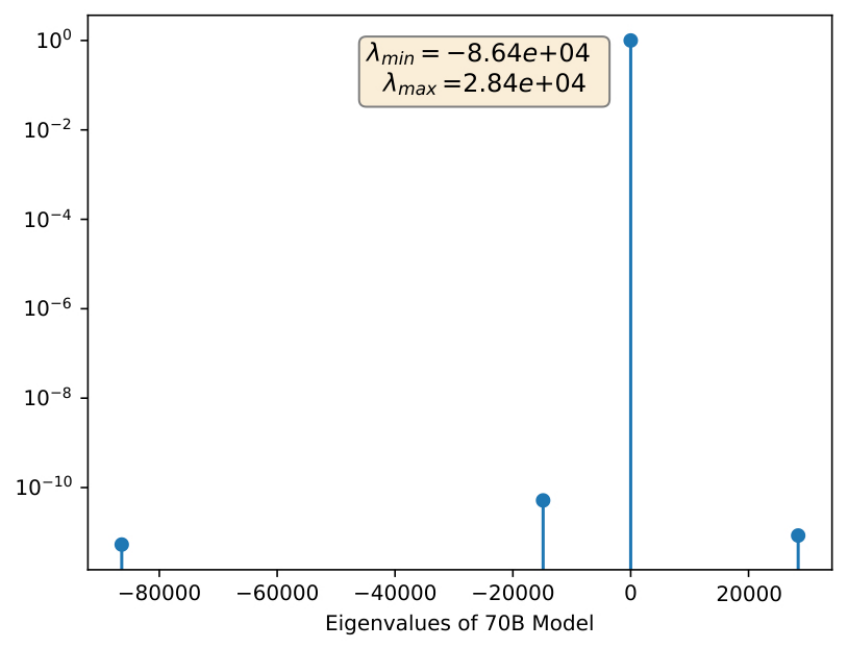}
		\caption{$1\%$ RealNews, 4  It}
		\label{fig:70b_1pc_realnews_4it}
	\end{subfigure}
	
	\caption{$70$ Bn DeepSeek Llama distillation: spectral widths under varying subsampling of Wikitext and RealNews.}
	\label{fig:70bn_subsample_row}
\end{figure}

We give here the first to our knowledge spectral density plots of the $70$bn Deepseek-Llama distilled frontier model in Figure \ref{fig:70bn_subsample_row}. We use a baseline of a $10$ lanczos iterations, except for in version in Figure \ref{fig:70b_1pc_realnews_4it}, where we run only $4$ on only $1\%$ of the realnews dataset, due to its huge computational cost. We see very interesting large negative outliers for the $70$bn model. This is strange because it indicates that there is a direction in the loss surface which if we were to descend we would heavily decrease the loss, an atpyical situation for a trained model. For frontier models that in theory generalise well to all data, as autoregressive models this is strange. We see that this is a general trend, which is robust to less aggressive (within our budget) subsampling, persists.

Given the ubiquity of the Pile dataset and the scraping wars to get as much data as possible to train large language models, a question of interest may be as to whether running the deepseek model on the pile dataset might give alternative results. Interestingly as shown in Figure \ref{subfig:pile70b} this is not the case.

\subsection{A note on rank degeneracy}
As the observant reader may have noticed, in all of our spectral plots, there seems to be a huge rank degeneracy around $0$. Unfortunately our spectral method, does not give us each of the eigenvalues for us to count. We only get a moment matched spectral approximation. Furthermore, the extent of our precision is limited as we do not store the vectors in memory (beyond a certain size) to allow us to keep orthonormality and stop the presence of ghosts. Furthermore even if we did, we would be limited by machine precision in our calculation. In this section we look at both issues in a bit more depth. Two examples of spectral results (known as the Ritz eigenvectors and weights) are given in Table \ref{tab:spectral_weights_sidebyside}. In one case, we see the large spectral peak at a number not very close to $0$ ($0.01$) but again note that only the moments of this spectral density must match the original, we could run this experiment again with a different seed and much more likely (from our evaluations at least) we would see something like in the second table. A huge rank degeneracy not at $0$ but very close $1e^{-6}$. Is this zero or not? Whilst a full rigorous analysis of this is beyond the scope of this work, consider the fact that for $32$ bit precision, we have one bit for the sign, eight bits for the exponent and tewnty-three for the mantissa, assuming that we represent zero exactly (usually the case in most frameworks) then machine epsilon is $2^{-23} = -23\times \log_{10}2 = 1.2 \times 10^{-7}$.  Interestingly the total weight to the other stems is in the region of $10^{-10}$. So one ten billionth of the model parameter directions determine the loss most strongly.
\begin{table}[htbp]
	\centering
	\caption{70B llama deepseek distilled spectral weights with 10 Lanczos iterations for wikitext\_tmat-2 (25\%) and realnewslike (1\%).}
	\label{tab:spectral_weights_sidebyside}
	%
	\begin{minipage}[t]{0.45\textwidth}
		\centering
		\begin{tabular}{rr}
			\toprule
			\multicolumn{2}{c}{Wikitext} \\
			\midrule
			Ritz Value & Weight \\
			\midrule
			-82221.5234 & 2.3228e-11 \\
			-81374.5859 & 1.2142e-11 \\
			-50410.7188 & 4.9422e-12 \\
			-36873.0273 & 4.7944e-11 \\
			-22338.5938 & 3.1632e-11 \\
			-7015.3286  & 1.9602e-10 \\
			0.0145      & 1.0000 \\
			4249.3477   & 3.7041e-10 \\
			11931.8496  & 2.0554e-11 \\
			21408.2852  & 8.0098e-12 \\
			\bottomrule
		\end{tabular}
	\end{minipage}
	\hfill
	\begin{minipage}[t]{0.45\textwidth}
		\centering
		\begin{tabular}{rr}
			\toprule
			\multicolumn{2}{c}{Realnewslike} \\
			\midrule
			Ritz Value & Weight \\
			\midrule
			-72159.5469 & 1.4208e-13 \\
			-71471.6719 & 1.2187e-15 \\
			-29396.1230 & 2.6713e-18 \\
			-11272.0820 & 2.4919e-13 \\
			-3630.3208  & 4.9708e-11 \\
			1.1277e-06  & 1.0000 \\
			1863.6674   & 3.0464e-10 \\
			8697.9453   & 1.2413e-11 \\
			18385.3047  & 3.4053e-12 \\
			23685.8652  & 1.0928e-12 \\
			\bottomrule
		\end{tabular}
	\end{minipage}
\end{table}
This puts us in an uncomfortable situation, we are more than $10\times$ away from machine precision. A more refined analysis might consider a Wigner model to the noise (additive noise) perturbation for the Hessian, however this would give us a semi circle of $\pm 2\sigma \sqrt{P}$ where $P$ could be in the billions and hence $\sigma$ would have to be a function of $P$. Further work could look a multiplicative noise model to settle the question as to whether the Hessian really is low rank, or just has very many nearly flat directions. From a practical perspective this is less important (we can still compress very flat directions with limited loss impact), from a theoretical perspective, quite flat and flat are very different.
\section{Banded Hessians}
In order to test for banded matrices, we in general have to go through a $\mathcal{O}(P^{2})$ calculation, which is infeasible. So we initially start with a simpler subproblem. A dense matrix with no zero values is trivially maximally banded. As such a matrix can only have a tight band if it is sparse, so we check for sparsity as a simpler condition instead. We can indeed check for sparsity of a particular column using our Hessian vector product. In this case we simply generate a random vector which has a random index chosen as $1$ and the rest zero. We could alternatively create the full Hessian column by column (and abruptly stop at the point of generation). 
\begin{figure}[h!]
	\centering
	\begin{subfigure}[b]{0.30\textwidth}
		\includegraphics[width=\textwidth]{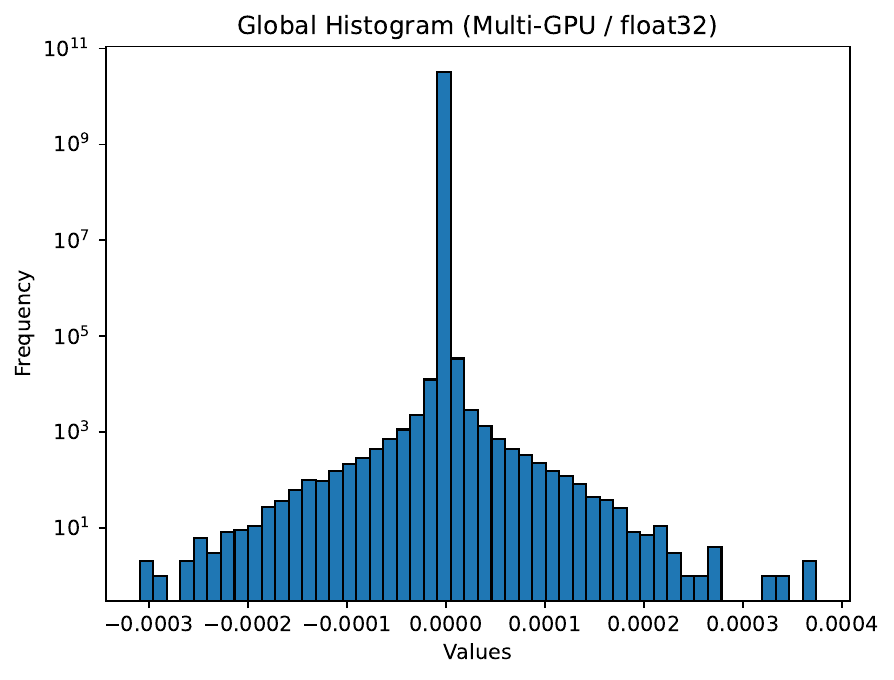}
		\caption{DeepSeek 32B (1 \%)}
		\label{fig:histogram-deepseek-32b-1pct}
	\end{subfigure}
	\hfill
	\begin{subfigure}[b]{0.30\textwidth}
		\includegraphics[width=\textwidth]{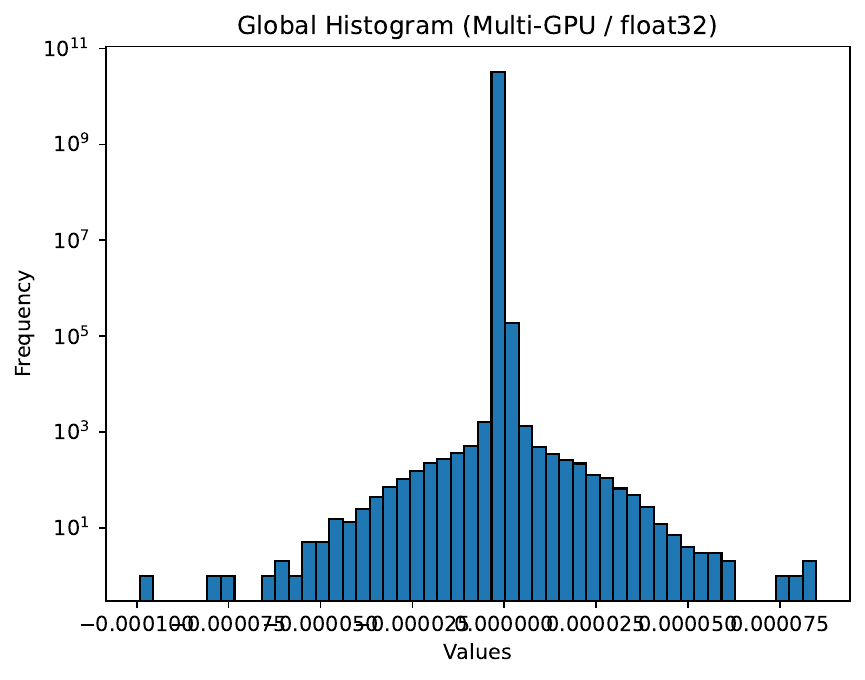}
		\caption{DeepSeek 32B (5 \%)}
		\label{fig:histogram-deepseek-32b-5pct}
	\end{subfigure}
	\hfill
	\begin{subfigure}[b]{0.30\textwidth}
		\includegraphics[width=\textwidth]{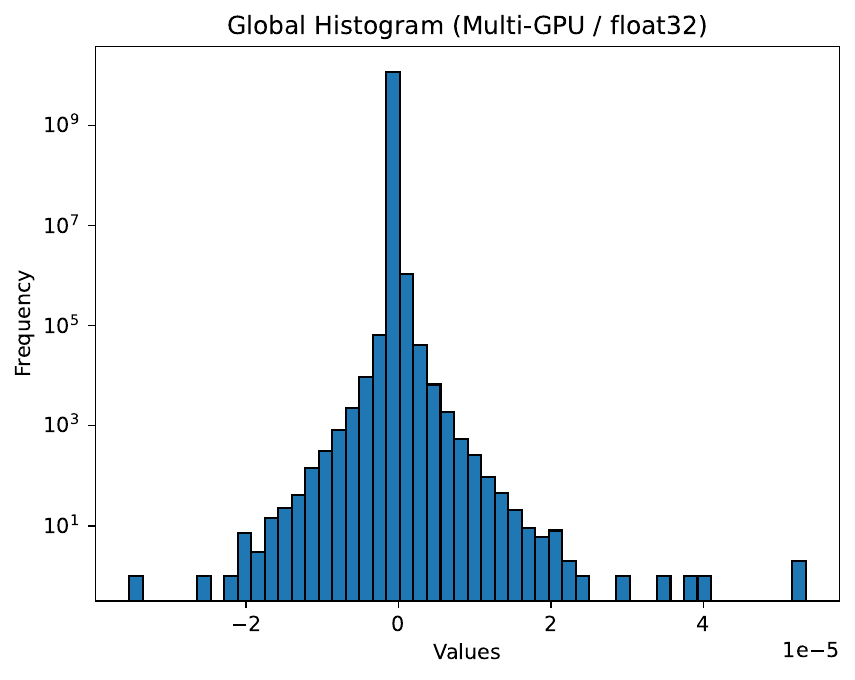}
		\caption{Pythia 12B (5 \%)}
		\label{fig:histogram-pythia-12b-5pct}
	\end{subfigure}
	
	\caption{50-bin absolute-magnitude histograms of Hessian-column elements for DeepSeek 32B (1 \% and 5 \%) and Pythia 12B (5 \%).}
	\label{fig:histogram_combined}
\end{figure}

We see that for both Deepseek $32$Bn in Figures \ref{fig:histogram-deepseek-32b-1pct},\ref{fig:histogram-deepseek-32b-5pct} and for Pythia in Figure \ref{fig:histogram-pythia-12b-5pct} where we plot the histogram of the absolute magnitude of the entries of the randomly chosen column vector that there is a large peak at $0$. Unsurprsingly with greater subsampling the values themselves broaden (variance and hence mean square value increases). However as shown in the extensive Tables \ref{tab:threeSideBySide}, we see that the number of elements smaller than machine epsilon is very nearly $100\%$. So there is evidence that the Hessian is sparse to within machine precision, which means that potentially within machine precision it could be banded. In general finding the minimum band is known to be NP hard. Certain heuristic methods like Reverse Cuthill-McKee require breadth first search and so are $\mathcal{O}P^{2}$ in time complexity and hence infeasible, so potentially a sampling based RCM method could be used to see what the band size could be capped at.

\begin{table}[t]
	\centering
	\tiny
	
	\setlength{\tabcolsep}{2pt}
	\caption{Fraction of Hessian-column elements below each threshold for different models and random seeds.}
	\label{tab:hessians-all}
	
	\subcaptionbox{Qwen-32B, 5 \%\label{tab:qwen32-5}}[.29\linewidth]{%
		\begin{tabular}{c|c}\hline
			Threshold & Fraction\\\hline
			1e$^{-12}$ & 0.0343\\
			1e$^{-11}$ & 0.0823\\
			1e$^{-10}$ & 0.2880\\
			1e$^{-9}$  & 0.8751\\
			1e$^{-8}$  & 0.9978\\
			1e$^{-7}$  & 0.9999\\
			1e$^{-6}$–1e$^{-1}$ & 1.0000\\\hline
			\multicolumn{2}{l}{\scriptsize Total elems = 32.8 B}
	\end{tabular}}%
	\hfill
	\subcaptionbox{Pythia-12B, 5 \%\label{tab:py12-5}}[.29\linewidth]{%
		\begin{tabular}{c|c}\hline
			Threshold & Fraction\\\hline
			1e$^{-12}$ & 0.0059\\
			1e$^{-11}$ & 0.0121\\
			1e$^{-10}$ & 0.0585\\
			1e$^{-9}$  & 0.3738\\
			1e$^{-8}$  & 0.9427\\
			1e$^{-7}$–1e$^{-1}$ & 1.0000\\\hline
			\multicolumn{2}{l}{\scriptsize Total elems = 11.8 B}
	\end{tabular}}%
	\hfill
	\subcaptionbox{Pythia-12B, 5 \% — five seeds\label{tab:py12-seeds}}[.39\linewidth]{%
		\centering
		\tiny
		\setlength{\tabcolsep}{1.5pt}
		\begin{tabular}{c|c|ccccc}\hline
			Idx & Thres. & F1 & F2 & F3 & F4 & F5\\\hline
			0 & 1e$^{-12}$ & 0.0067 & 0.0076 & 0.0087 & 0.0069 & 0.0095\\
			1 & 1e$^{-11}$ & 0.0158 & 0.0213 & 0.0278 & 0.0167 & 0.0373\\
			2 & 1e$^{-10}$ & 0.0833 & 0.1246 & 0.1693 & 0.0912 & 0.2458\\
			3 & 1e$^{-9}$  & 0.5035 & 0.6633 & 0.8241 & 0.5463 & 0.8834\\
			4 & 1e$^{-8}$  & 0.9921 & 0.9983 & 0.9993 & 0.9929 & 0.9977\\
			5 & 1e$^{-7}$  & 0.9998 & 0.9999 & 1.0000 & 0.9999 & 1.0000\\
			6 & 1e$^{-6}$–1e$^{-1}$ & \multicolumn{5}{c}{1.0000}\\\hline
			\multicolumn{7}{l}{\scriptsize 5 iterations per seed}
	\end{tabular}}
\end{table}

\section{Compute resources.}
All Hessian–vector products (HVPs) were executed on a single server (type \#19983985) with  
\textbf{8 × NVIDIA H200 140 GB PCIe 5.0 GPUs} (PCIe 5.0 ×16, 52.8 GB s\(^{-1}\) host–device; no NVLink),  
dual Intel \textsc{Xeon} Platinum 8568Y+ CPUs (192 logical cores), and 2 TB DDR5 RAM.  
Local storage: 2 × Dell NVMe PM1733a 3.84 TB SSDs in RAID-0 (12.4 GB s\(^{-1}\) sequential reads).  
Software: Ubuntu 22.04, CUDA 12.8, cuDNN 9.0, PyTorch 2.3.

\paragraph{Compute accounting.}
\begin{itemize}
	\item \textbf{HVPs.} Each run took 3 h wall-clock; \(3 \text{h}\times8\text{ GPUs}=24\) GPU-h.  
	With 20 HVPs: \(20\times24 = 480\) GPU-h.  
	Applying a 1.5× overhead for checkpoints, failed starts, and analysis yields  
	\(480\times1.5 = 720\) GPU-h.
	\item \textbf{Smaller experiments.} Hyper-parameter sweeps, ablations, and sanity-check
	trainings were of comparable scale; we conservatively allocate an
	\emph{additional} \(\approx720\) GPU-h.
\end{itemize}

\[
\boxed{\text{Total compute} \;=\; 720 + 720 = 1\,440\ \text{GPU-hours}
	\;=\; 60\ \text{GPU-days on H200 PCIe}}
\]

Using CodeCarbon v2.3 with the default grid mix,
this corresponds to \(\approx66\;\text{kg CO\(_2\)e}\).

\section{Conclusion}\label{sec:conclusion}

In this paper, we introduce \textbf{HessFormer} a method and software to calculate the empirical spectral density of foundation models that works within the popular Pytorch and Huggingface ecosystem. We validate the accuracy of the method against by running on smaller models that fit on a single GPU or where full orthogonalisation is possible and show minimal deviation from these idealised cases. We investigate the Spectra of Deepseek-$70$bn and comment on key aspects therein. Whilst our distributed stochastic lanczos quadrature is state of the art in terms of investigating the largest deep neural networks to date, an obvious limitation is that we are inherrently limited by the estimated spectral moments (using stochastic trace estimation) to the order of the number of Lanczos iterations, which again are unable to go beyond $m=10$ due to loss of orthogonality, exemplified by the appearance of ghost eigenvalues in our experiments. Our method is compute in-efficient relative to FSDP and at the time of writing the state of the art known open-source models, have over $700$B parameters. Therefore our current experimental contribution is an order of magnitude off allowing curvature calculations for SoTA models. We also do not repeat the experiments with multiple seeds (data or random vector) and since we do not run many random vectors we do not run an average. However due to the self averaging property of large random vectors and matrices, we do not expect this to have major impact on the results of our research.

\section*{Broader Impact}
Public access to reliable Hessians at the billion-parameter scale enables safer optimisation decisions, principled influence estimation, energy-efficient model compression and real-time curvature-based monitoring, thereby supporting both performance and alignment goals.  At the same time, detailed curvature information could facilitate model extraction, jailbreak attacks and rapid miniaturisation of powerful systems for malicious use.  To balance these considerations, HessFormer is released under a research-only licence, with exploit-specific details withheld and an explicit call for users to integrate curvature analyses into existing responsible-AI protocols while remaining vigilant to emerging dual-use scenarios. Our work, as all foundation model scale gradient work brings with it an ever increasing and unsustainable energy and environmental impact. However, increased faithfulness to the full Hessian may impact the ability to compress models (reducing inference energy) and reduce per flop optimisation (reducing training energy) and so we are optimistic that this research could pave the way for less and not more energy useage.

\bibliographystyle{plainnat}
\bibliography{bib}

\end{document}

%% file: maths_commands.tex
\definecolor{RED}{rgb}{0.7,0,0}
\definecolor{BLUE}{rgb}{0,0,0.69}
\definecolor{GREEN}{rgb}{0,0.6,0}
\definecolor{PURPLE}{rgb}{0.69,0,0.8}
\definecolor{ORANGE}{RGB}{255,103,0}
\definecolor{BROWN}{RGB}{100,20,45}

